\documentclass[copyright,creativecommons]{eptcs}
\providecommand{\event}{ICLP 2025} 

\usepackage{iftex}

\ifpdf
  \usepackage{underscore}         
  \usepackage[T1]{fontenc}        
\else
  \usepackage{breakurl}           
\fi

\newtheorem{example}{Example}[section]

\usepackage{todonotes}
\usepackage{multirow}

\newboolean{includeMemo}

\setboolean{includeMemo}{false}

\newcommand{\memoside}[1]{\ifthenelse{\boolean{includeMemo}}{\todo[caption={},color=green!20!]{{\footnotesize #1}}}}
\newcommand{\memo}[1]{\ifthenelse{\boolean{includeMemo}}{\todo[inline,caption={},color=green!20!]{#1}}}
\newcommand{\highlightbox}[1]{\ifthenelse{\boolean{includeMemo}}{\todo[inline,caption={},color=gray!20!]{#1}}}

\usepackage{booktabs}

\usepackage{newfloat}

\usepackage{listings}
\lstdefinelanguage{clingo}{ 
  keywordstyle=[1]\usefont{OT1}{cmtt}{m}{n},%
  keywordstyle=[2]\textbf,%
  keywordstyle=[3]\usefont{OT1}{cmtt}{m}{n},
  alsoletter={\#,\&},%
  keywords=[1]{not},%
  keywords=[2]{\#const,\#show,\#minimize,\#maximize,\#base,\#theory,\#count,\#external,\#program,\#script,\#end,\#heuristic,\#edge,\#project,\#show},%
  keywords=[3]{&,&dom,&sum,&diff,&show,&minimize},%
  morecomment=[l]{\#\ },%
  morecomment=[l]{\%\ },%
  commentstyle={\color{darkgray}}%
}%
\usepackage{wrapfig}
\usepackage{picinpar} 

\newcommand{\RED}[1]{\textcolor{red}{#1}}

\title{An LLM + ASP Workflow for Joint Entity-Relation Extraction}
\author{Trang Tran 
\institute{New Mexico State University \\ New Mexico, USA}
\email{ttran@nmsu.edu}
\and
Trung Hoang Le
\institute{New Mexico State University \\ New Mexico, USA}
\email{trungle@nmsu.edu}
\and
Huiping Cao
\institute{New Mexico State University\\
New Mexico, USA}
\email{hcao@nmsu.edu}
\and 
Tran Cao Son
\institute{New Mexico State University \\ New Mexico, USA}
\email{stran@nmsu.edu}
}
\def\titlerunning{LLM + ASP Workflow}
\def\authorrunning{T. Tran, T. H. Le, H. Cao, T. C. Son}
\begin{document}

\makeatletter

\maketitle

\begin{abstract}
Joint entity-relation extraction (JERE) identifies both entities and their relationships simultaneously. Traditional machine-learning based approaches to performing this task require a large corpus of annotated data and lack the ability to easily incorporate domain specific information in the construction of the model. 
Therefore, creating a model for JERE is often labor intensive, time consuming, and elaboration intolerant.
In this paper, we propose harnessing the capabilities of generative pretrained large language models (LLMs) and the knowledge representation and reasoning capabilities of Answer Set Programming (ASP) to perform JERE. 
We present a generic workflow for JERE using LLMs and ASP. 
The workflow is generic in the sense that it can be applied for JERE in any domain. 
It takes advantage of LLM's capability in natural language understanding in that it works directly with unannotated text. 
It exploits the elaboration tolerant feature of ASP in that 
no modification of its core program is required when  additional 
domain specific knowledge, in the form of type specifications, is found and needs to be used.  
We demonstrate the usefulness of the proposed workflow through experiments with limited training data on three well-known benchmarks for JERE. 
The results of our experiments show that the LLM + ASP workflow 
is better than state-of-the-art JERE systems in several categories with only 10\% of training data.
It is able to achieve a 2.5 times (35\% over 15\%) improvement in the Relation Extraction task for the SciERC corpus, one of the most difficult benchmarks. 
\end{abstract}

\section{Introduction}\label{sec:intro}
 
Named Entity Recognition (NER) and Relationship Extraction (RE) are classification tasks in Natural Language Processing (NLP) focused on identifying and labeling entities and relationships from unstructured data into predefined categories as discussed by \cite{IEEE:journals/deeplearning/Li}. Both tasks are useful in information extraction, knowledge graph creation, and question answering~(\cite{survey:journals/RE/,IEEE:journals/deeplearning/Li}). When both tasks are performed simultaneously by the same model, it is known as joint entity-relation extraction (JERE) as defined by~\cite{zheng201759}.  
It is well-known that JERE is much harder than NER or ER. 
Traditionally, supervised models are most effective when trained on large sets of domain specific annotated data for NER and RE tasks~(see, e.g., \cite{llmNER:villenaa}). 

In recent years, Large Language Models (LLMs) 
such as Generative Pretrained Transformer (GPT) 
have been used in information extraction tasks with techniques such as instruction tuning (\cite{Wang2023InstructUIEMI}), transforming sequence labeling tasks into generation tasks (\cite{wang2023gptnernamedentityrecognition}), and augmenting datasets for finetuning (\cite{santoso2024lowresourcener}). 
To the best of our knowledge, 
LLMs have not been used specifically for the JERE task. Nevertheless, \cite{bonfligi2024103003} has shown that fine-tuning a GPT can enhance results in NER tasks with the improvements depending on the amount of data used to fine-tune. 
 
This paper introduces a novel approach that combines LLMs with logic programming under answer set semantics (ASP), introduced by \cite{GelfondL90}, to jointly identify entities and their relationships from unstructured text. This design leverages the vast knowledge base embedded in GPT, which has been pretrained on billions of data points across various domains. While GPT's initial capabilities are impressive, they are still prone to producing hallucinations, which are falsified information presented as fact about real-world subjects~as discussed by \cite{tonmoy2024hallucination}. 
In this paper, we propose using ASP and domain specific knowledge, whenever it is available, to mitigate false predictions generated by the LLM. 

The main contributions of this work are as follows. 
\begin{list}{$\bullet$}{\itemsep=0pt \parsep=0pt \topsep=1pt \leftmargin=10pt}
    \item A workflow that is elaboration tolerant by exploiting GPT's corpus and broad applicability along with ASP's flexibility for use in JERE tasks. It is a simple, but effective, workflow that shows how symbolic knowledge representation can further improve upon generative outputs from LLMs. 
    \item A modular prompt template that can be used for JERE tasks across domains. 
    \item An experimental evaluation demonstrating the superiority of the proposed approach compared to two state-of-the-art methods, using three commonly used benchmark datasets for JERE.
\end{list}
The next section presents the necessary background and related works in JERE and ASP. 
Section~\ref{sec:methods} details our approach. Section ~\ref{sec:experiment} describes our experiments and their results.  
We conclude the paper in Section~\ref{sec:conclusion}.

\section{Background and Related Works}\label{sec:related}
 
An ASP program consists of rules of the form ``$head \leftarrow body$'' where $head$ is an atom and $body$ is a conjunction of atoms or default negations of atoms in a first order language. Intuitively, a rule states that if the $body$ is true then the $head$ must be true. 
Answer set semantics of a logic program is define by 
\cite{GelfondL90} and can be computed efficiently using 
answer set solvers such as \texttt{clingo}\footnote{
\url{https://potassco.org}
}. 
In this paper, we employ ASP with extended syntax such as choice atoms, aggregate atoms, and constraints that has become the standard of logic programming language and implemented in most answer set solvers.  
It is worth noticing that, recently, ASP has been used to enhance the logical reasoning and accuracy of GPT outputs in code generation tasks and spatial reasoning as discussed in  \cite{kalyanpur2024llmarcenhancingllmsautomated} and \cite{wang2024dspybasedneuralsymbolicpipelineenhance}. 
The consistency checking step in this work is inspired by the system ASPER (see below). 

The literature on NER, ER, and JERE spans a wide range of methods, from traditional rule-based approaches to more recent machine learning and deep learning techniques. The surveys by \cite{IEEE:journals/deeplearning/Li} and \cite{survey:journals/RE/} mainly focus on NER. Early work in NER, ER, and JERE often relied on handcrafted rules and annotators to identify entities and relationships, which limited scalability and accuracy. With the rise of supervised learning, models like Conditional Random Fields (CRFs) and Support Vector Machines (SVMs) were introduced, allowing for more flexible and data-driven extraction. 
Deep learning approaches, especially those based on transformer architectures (e.g., BERT, RoBERTa), have significantly advanced JERE by leveraging pretrained contextual embeddings. These models have shown superior performance in a variety of domains, including biomedical text mining, legal document analysis, and social media content.  Most recently, prompt engineering frameworks have been implemented to take advantage of pretrained large language models. InstructUIE\cite{Wang2023InstructUIEMI} uses multi-task instruction and fine-tuning to identify named entities (without classifying their types) and to extract relationships between entities separately, rather than jointly detecting entities, their types, and their relationships. 
RIIT-IE\cite{geng2024} attempts to distill noise from true positives when detecting entities and their relationships, using iterative and hierarchical prompt engineering. Among prompt-only methods, its framework achieves the best performance we have seen. However, RIIT-IE employs significantly more complex prompting techniques. It uses a modular system where data pass through multiple layers, with different prompts applied at each layer to progressively narrow down the correct answers. 

Despite the advances, 
challenges still remain, such as handling ambiguous entities, identifying novel relationships, and extracting information from noisy or unstructured data. As a result, ongoing research is focused on enhancing model generalization, developing domain-specific models, and incorporating external knowledge sources to further improve the accuracy and robustness of joint entity relationship extraction. 

To the best of our knowledge, the system ASPER, developed by~\cite{DBLP:journals/tplp/LeCS23}, performs better than state-of-the-art JERE systems when it was introduced. It is shown in that paper that domain specific knowledge can be exploited effectively in reducing the amount of training data and in increasing the model performance. 
ASPER employs ASP to improve the learning process of neural network models. 
ITER, the most recent introduction to the JERE landscape by~\cite{hennen-etal-2024-iter}, 
is currently the best system for JERE. It is an encoder-only, transformer-based model. 
ITER's performance, however, depends on the amount of data used in its training.

\section{A {Lightweight} LLM + ASP Approach}\label{sec:methods}
\subsection{Overview of the Proposed Method}

\begin{figwindow}[2,r,%
{
\includegraphics[width=0.7\textwidth]{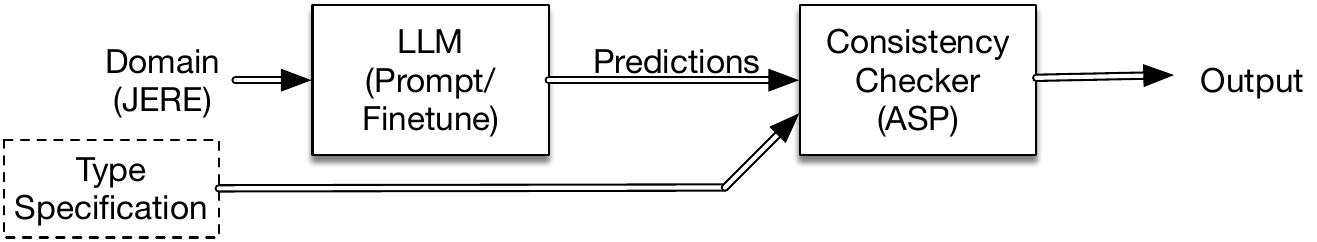} 
},
{
LLM + ASP Workflow for JERE
\label{fig:workflow}}] 
\noindent
We propose a lightweight workflow to conduct effective JERE. 
The framework consists of two main components (Figure~\ref{fig:workflow}): (\emph{i}) a generic prompt template for JERE, given the domain and annotation guideline; and (\emph{ii}) a consistency checker that is written in ASP. The template aims at asking a LLM, GPT or Gemini in our study, to extract entities and relations from the domain. Use of a retrained LLM can take advantage of the knowledge that is learned in the model and save time on training another new machine learning model. At the same time, it is well known that LLMs such as GPT produce hallucinations~(see, e.g., (\cite{perkovic2024})). This means that the entities and relationships returned from an LLM model may have both false positives and false negatives. To help improve the quality of LLM output, we design a novel strategy to verify the consistency of the output and eliminate inconsistent outputs. More concretely, in step (\emph{ii}),  
the output of an LLM model 
is then provided to an ASP solver together with available domain-specific knowledge, called \emph{type specification}, for consistency checking.
\end{figwindow}  
We next detail the design of the prompt template (Section~\ref{sec:promttemplate}) and the ASP program for consistency checking (Section~\ref{sec:asp}).

\subsection{Pre-study}
\label{sec:prestudy}

To create an effective and generic template for JERE, we conducted a comprehensive study on the 
state-of-the-art prompting techniques for entity-relation extraction tasks. Techniques such as In-Context-Learning, Chain-Of-Thought, zero-shot, and few-shot prompting 
(\cite{schulhoff2024promptreportsystematicsurvey}) were tested  to see which (or a combination of them)
would yield good results with respect to our task. We employed GPT-3.5 and used the ConLL04 as a sample domain to conduct this preliminary study. 
The experiments showed that the following four techniques in combination resulted in a 6\% increase in the F1-macro score of both entity and relationship extraction tasks without fine-tuning. They are: 
\begin{list}{$\bullet$}{\topsep=1pt \itemsep=0pt \leftmargin=10pt}
    \item Giving GPT specific context by clearly defining a domain and the role it will take on.
    \item The use of one-shot prompting by including one example.
    \item Addition of constraints and definitions for what is considered an entity or relationship in the confines of our dataset. 
    \item Answer engineering in which we defined the output key specifications in JSON format. 
\end{list}
We used the prompt building techniques we learned through this pre-study to inform our prompt engineering in the next section.

\memo{Son: rewording as above \\ 
\RED{(Add 1-3 sentences to justify why trying these 13 variations.)} \textcolor{teal}{With each iteration, we compared the results with previous iterations of the prompt - stopping only when the threshold for improvement reached a peak.} In the end, we compared 13 variations of prompts on GPT-3.5 using the CoNLL04
dataset~\cite{SPERT2020} to see which techniques yielded the best results.

Prior to our experiments, we conducted a comprehensive study on the 
state-of-the-art prompting techniques for entity-relation extraction tasks. Techniques such as In-Context-Learning \cite{schulhoff2024promptreportsystematicsurvey}, Chain-Of-Thought, zero-shot, and few-shot prompting were tested iteratively to see which (or a combination of which) would yield good results. \RED{(Add 1-3 sentences to justify why trying these 13 variations.)} \textcolor{teal}{With each iteration, we compared the results with previous iterations of the prompt - stopping only when the threshold for improvement reached a peak.} In the end, we compared 13 variations of prompts on GPT-3.5 using the CoNLL04
dataset~\cite{SPERT2020} to see which techniques yielded the best results.

The experiments showed that the following four techniques in combination resulted in a 6\% increase in the F1-macro score of both entity and relationship extraction tasks without fine-tuning. They are: 
\begin{itemize}
    \item Giving GPT specific context by clearly defining a domain and the role it will take on.
    \item The use of one-shot prompting and the inclusion of one example
    \item Addition of constraints and definitions for what is considered an entity or relationship in the confines of our dataset. 
    \item Answer Engineering by defining the output key specifications in JSON format 
\end{itemize}
We used the prompt building techniques we learned through this pre-study to inform our prompt engineering in the next section. 
} 

\subsection{Prompt Engineering} 
\label{sec:promttemplate}

For the JERE task, a prompt template needs to define the domain, experience, context, output keys, and one example. Since our goal is to create a JERE system that can work with arbitrary domains, we create 
a generic base template for the JERE task that can be augmented with domain-specific information. In this sense, our template is similar to the modular template used in PromptNER, introduced by  \cite{ashok2023promptnerpromptingnamedentity}.

A {\em Domain} is a general term to narrow the field in which the LLM agent is asked to focus. {\em Experience} refers to how much experience the GPT agent has within the given domain. The {\em Context} includes general definitions for what is considered an entity or relationship, the types and how to annotate the text for the specified entity and relationship types. It is derived directly from the annotation guidelines for each dataset. The domain, experience, and context are assigned in the system prompt. This gives the GPT agent background knowledge of the task and domain. 

{\em Output Keys} refer to the specific keys used for evaluation. The output keys and one example are stated in the user prompt and give more specificity to the LLM agent. All of the user-defined categories above are informed by the annotation guides supplied by each dataset.

\begin{example}
\label{ex:template}
Below is an example of how a dataset has been broken down into the different categories and the full prompt: 

\begin{itemize}
	\item \textbf{Domain:} “journalism and news”
	\item \textbf{Experience:} 
    ``You have an M.Sc. degree in linguistics and substantial background working to annotate entities and relationships using your knowledge of syntax and semantics."
	\item \textbf{Context:} 
    “Only classify entity types as either location, organization, people, or other. Output ‘Loc' for location, ‘Peop' for people, ‘Org' for organization and ‘Other' for other. Only classify relationship types as either organization based in, located in, live in, work for, or kill. Output ‘OrgBased\_In' type for organization based in, ‘Located\_In' for located in, ‘Live\_In' for live in, ‘Work\_For' for work for, and ‘Kill' for kill."
	\item \textbf{Output Key:} 
    “entities”: [“entity”:, “type”:], “relationships”: [“subject”:, “object":, “type":]
	\item \textbf{Example:}
    ``Input: ``Andrew Jackson, born March 15, 1767, in Waxhaw settlement.", \\ 
    Output:\{``Entities":[\{``Entity": ``Andrew Jackson", ``Type":``Peop"\},\{``Entity": ``March", ``Type": ``Other"\},\{``Entity": ``Waxhaw", ``Type": ``Loc"\}], “Relationships":[\{“Subject": “Andrew Jackson",“Object": “Waxhaw", “Type": “Live\_In"\}\}"     
\end{itemize}
\end{example}

\subsubsection{Base Prompt Template}
Our prompt template consists of two components, system and user. They are defined as follows.

\noindent
\textbf{System:}

	“You are a natural language processing researcher working in the \{DOMAIN\} domain. \{EXPERIENCE\} Your job is to extract entities from the excerpts of texts given. In this domain, an entity is an object, set of objects or abstract notion in the world that has its own independent existence. Entities specify pieces of information or objects within a text that carry particular significance. In your work, you will only extract specific types of entities and relationships. The types of entities and relationships are defined here. \{CONTEXT\}”

\noindent
\textbf{User:} 

“Give me the entities from the following text. Do not include any explanations, only provide 
RFC8259 compliant JSON response without deviation. Do not include ‘$\backslash$n' (\texttt{newline}) in the output. The keys for the output JSON should be \{OUTPUT\_KEYS\}. Do not use any other keys for the JSON response. 
Ensure that you are outputting the entire entity and its type. 
Here is one example: \{EXAMPLE\}
Evaluate this text: \{TEXT\}” 

As we have seen with GPT's, the more specific a prompt is, the better the results, but it is time consuming to consider how to engineer a prompt for every situation. This template for both system and user prompts allowed us to generalize the task even when the datasets are not related. We must still determine what appropriate information should be included in a prompt, but the base template gives us guidance on what type of information is relevant and needed.


\subsection{Consistency Checking Using Answer Set Programming}
\label{sec:asp}

To eliminate potential false predictions from the output of the LLM, we propose a verifying step, termed as \emph{consistency checking}. This step takes advantage of the facts that the LLM's output is essentially a collection of atoms, and thus, can be easily manipulated via rules. The idea of utilizing ASP to conduct consistency check originated from ASPER~\cite{DBLP:journals/tplp/LeCS23}. However, the available data structure in this work is completely different from that used in ASPER and therefore, the code in this work is different from that used in ASPER and, we believe, is much easier to understand.

\subsubsection{ASP Program for Consistency Checking}

We  describe the program, denoted by $\Pi_C$, that is the main ingredient of the consistency checking step. This program takes the output of the GPT encoded as a collection of facts of the forms 
\begin{list}{$\bullet$}{ \itemsep=1pt \parsep=0pt \topsep=0pt \leftmargin=10pt}
    \item \lstinline{atom(ent(S,E,T))}: 
    $E$ is an entity of the type $T$ in the sentence $S$; and 
    \item 
    \lstinline{atom(rel(S,E,F,R))}: 
    relation of type $R$ between entities $E$ and $F$ in the sentence $S$.
\end{list}
Optional inputs of the program include 
\begin{list}{$\bullet$}{\itemsep=1pt \parsep=0pt \topsep=0pt \leftmargin=10pt} 
    \item Type specification of the form  \lstinline{type_def(R,T,V)}: 
    relation of type $R$ is between entities of the types $T$ and $V$; 
    \item Ground truth of the form 
    \lstinline{ent(S,E,T)} and \\
    \lstinline{rel(S,E,F,R)} whose meaning is similar to that of 
    \lstinline{atom(ent(_))} and 
    \lstinline{atom(rel(_))}, respectively.
\end{list}
The program defines the following predicates: 
\begin{list}{$\bullet$}{\itemsep=1pt \parsep=0pt \topsep=0pt} 
    \item \lstinline{false_declaration(S,E,F,R)}: 
    at least one of the entities, $E$ or $F$, of the relation $R$ does not appear in the entity list;
    \item \lstinline{ok_type(S,E,F,R)}: the type of the relation $R$ between $E$ and $F$ matches its specification; and 
    \item \lstinline{has_declaration(R)}: the type of the relation $R$ is specified.
\end{list}

The predicate \lstinline{false_declaration} encodes relations that are \emph{inconsistent} with the set of entities while  \lstinline{ok_type} reports relations that are \emph{consistent} with the type specification. 
These predicates are defined by the following rules: 

\begin{lstlisting}[language=clingo,caption=ASP Program for Consistency Checking][t]
false_declaration(S,E,F,R):-  atom(rel(S,E,F,R)), 
    1{not atom(ent(S,E,_)); not atom(ent(S,F,_))}.
has_declaration(R) :- type_def(R, _, _). 
ok_type(S,E,F,R):-atom(rel(S,E,F,R)),atom(ent(S,E,T)),atom(ent(S,F,V)),
    1{type_def(R,T,V); not has_declaration(R)}. 
\end{lstlisting}

We denote the above set of rules by $\Pi_C^1$. 
The first rule (Lines 1-2) defines when a relation has false declaration.  
The atom 
\lstinline|1{not atom(ent(S,E,_)); not atom(ent(S,E',_))}|
(Line 2)
 indicates that at least one of the atoms 
\lstinline{atom(ent(S,E,_))} or 
\lstinline{atom(ent(S,F,_))} is not contained in the output of the model, i.e., 
either $\mathtt{E}$ or $\mathtt{F}$ was not detected as  
an entity by the LLM.
The rule defining \lstinline{ok_type(S,E,F,R)} (Lines 4--5) says that the type of the relation ($\mathtt{R}$) is appropriate given the type specification or the type 
of the relation $\mathtt{R}$ is unspecified.
This allows for the program to be used with or without type specification. Line 3 is used to indicate that domain-specific information is available.

Given the model output $O$ and set of type specification atoms $D$, it is easy to see that the program $\Pi_C^1 \cup O \cup D$ has a unique answer set $O \cup D \cup W$ where $W$ is a collection of atoms of the form 
\lstinline{false_declaration(s,e,f,r)}, 
\lstinline{has_declaration(r)}, and 
{\lstinline{ok_type(s,e,f,r)}}. Note that if $D = \emptyset$, i.e., type specification is not available, then all RE predictions have the correct type, and thus, are acceptable.   
We consider \lstinline{rel(s,e,f,r)} as invalid if 
$W$ contains \lstinline{false_declaration(s,e,f,r)} or 
does not contain \lstinline{ok_type(s,e,f,r)} and remove it from the output of the model.

The next set of rules can be used for computing the various components needed for computing the F1-scores (macro-F1 and micro-F1). 
When the ground truth is not provided, these rules are not activated and will not change the content of the answer set of $\Pi_C^1 \cup O \cup D$.
In the code, \lstinline{#count} refers to the aggregate counting the number of elements in a set specified between the brackets \{ and \}. 
The rules defining the predicates
\lstinline{r_true_pos/4} (Lines 7--8)
and 
\lstinline{r_false_pos/4} (Lines 9--11)
remove predictions with incorrect type  or false declaration from consideration. The meaning of the other predicates is easily understood and is therefore omitted for brevity.  

\medskip

\begin{lstlisting}[language=clingo,caption=Computing True/False Positive/Negative and F1-score][t]
in_set(S):-atom(ent(S, _, _)). in_set(S):-atom(rel(S, _, _, _)).
r_true_pos(S,E,F,R):- atom(rel(S,E,F,R)), 
    ok_type(S,E,F,R),rel(S,E,F,R).
r_false_pos(S,E,F,R):- atom(rel(S,E,F,R)), ok_type(S,E,F,R), 
    not false_declaration(S,E,F,R), not rel(S,E,F,R). 
r_false_neg(S,E,F,R):- rel(S,E,F,R), in_set(S), not atom(rel(S,E,F,R)).   
r_true_p_cnt(C,T):-type_of_r(T),C=#count{S,E,F:r_true_pos(S,E,F,T)}.
r_false_p_cnt(C,T):-type_of_r(T),C=#count{S,E,F:r_false_pos(S,E,F,T)}.
r_false_n_cnt(C,T):-type_of_r(T),C=#count{S,E,F:r_false_neg(S,E,F,T)}.
e_true_pos(S,E,T):-ent(S,E,T), atom(ent(S,E,T)). 
e_false_pos(S,E,T):- atom(ent(S,E,T)),not ent(S,E,T). 
e_false_neg(S,E,T):-ent(S,E,T),in_set(S), not atom(ent(S,E,T)).
true_p_cnt(C,T):-type_of_ent(T),C=#count{S,E:e_true_pos(S,E,T)}.
false_p_cnt(C,T):-type_of_ent(T),C=#count{S,E:e_false_pos(S,E,T)}.
false_n_cnt(C,T):-type_of_ent(T),C=#count{S,E:e_false_neg(S,E,T)}.
\end{lstlisting}

\section{Experimental Evaluation}\label{sec:experiment}

\subsection{Experimental Settings}

Python code was implemented using Python 3.10 and OpenAI SDK version 1.57.0 and performed on a MacBook Pro with an Apple M3 Max chip. The fine-tuning and JERE tasks were run on OpenAI's servers and call the gpt-4o-2024-08-06 model \cite{openai}, referred to as \textbf{GPT} from now on. Specifically, we use the Batch and Fine-Tuning APIs from OpenAI. For the ensemble experiment, we use Google's Gemini  Flash 1.5 \cite{gemini}, referred to as \textbf{Gemini}, and the google-generativeai API version 0.8.3. 
The ASP solver is \lstinline{clingo} 5.4.0 ~\cite{gekakasc14b}. 
Source code and execution instruction related to the project can be found at the github~  
\cite{github}.

\smallskip\noindent
\emph{Data.} 
Our work focuses on joint entity and relation extraction (JERE) identifying entities with their types and predicting relations between them within a single sentence.  
Therefore, we select the following benchmarks for our experiment:

\begin{list}{$\bullet$}{\topsep=1pt \itemsep=0pt \leftmargin=10pt}
\item 
{\bf CoNLL04}~(\cite{SPERT2020,DBLP:conf/conll/RothY04,DBLP:conf/coling/GuptaSA16,wang2020two}): This dataset contains a total of 1,437 sentences retrieved from newspaper clippings and resides in the `news and journalism' domain. It differentiates between 4 types of entities (\lstinline{people, organization, location}, and \lstinline{other}) and 5 types of relationships (\lstinline{live_in,} \lstinline{ located_in, kill, orgbased_in,} and \lstinline{work_for}).
\item 
{\bf SciERC}~(\cite{SPERT2020,DBLP:conf/emnlp/LuanHOH18}): This dataset contains 2,412 sentences from scientific abstracts and differentiates between 6 types of entities (\lstinline{task, method, metric, material, otherScientifcTerm,} and \lstinline{generic}) and 7 relationships (\lstinline{compare, part-of, conjunction, evaluate-for, feature-of,} and \lstinline{used-for, hyponym-of}). 
\item 
{\bf ADE} \cite{ADE:journals/bi/Gurulingappa}: it contains 4,272 annotated documents from the `health and drug' domain and differentiates between 2 types of entities (\lstinline{drug} and \lstinline{adverse-effect}) and one relationship (\lstinline{adverse-effect}). 
\end{list}

We note that there are other well-known benchmarks such as the TACRED, REFinD, SemEval-2010 Task 8 and DocRED datasets\footnote{\url{https://nlp.stanford.edu/projects/tacred/}, \url{https://refind-re.github.io}, \url{https://arxiv.org/pdf/1911.10422},  \url{https://arxiv.org/pdf/1906.06127}} that were used by some entity/entity-relation extraction systems. However, TACRED, SemEval-2010 and REFinD are designed to annotate entity pairs and their relationships within individual sentences, and hence, they may overlook other entities in the sentence, limiting their suitability for full entity extraction. DocRED consists of multi-sentence instances where the same entity can appear in different forms and locations within an instance, requiring entity resolution before applying the JERE task.
 
We note that there are domains rich in type specification such as the CoNLL04 domain.
For example, the following relationships between types of the relations and entities were introduced by ~\cite{DBLP:journals/tplp/LeCS23}:  

\smallskip

\begin{lstlisting}[language=clingo,caption=Type Specification CoNLL04][t]
type_def("located_in","loc","loc"). type_def("live_in","peop","loc").
type_def("orgbased_in","org","loc").type_def("work_for","peop","org").
type_def("kill", "peop", "peop").
\end{lstlisting}
For the SciERC dataset, we derive a set of type specifications for this domain given the set of entities. 
Given the intuitive meaning of the entity types in the domain, we consider the following possible combinations of the \lstinline{part-of} relation:  

\smallskip \noindent
\begin{lstlisting}[language=clingo,caption=Type Specification SciErc][t]
type_def("part-of","task","task").
type_def("part-of","generic","generic").
type_def("part-of","material","material").
type_def("part-of","otherscientificterm","otherscientificterm").
type_def("part-of","metric","metric").
type_def("part-of","method","method").
type_def("part-of","otherscientificterm","method").
type_def("part-of","generic","method").
type_def("part-of","method","generic").
type_def("part-of","task","method").
\end{lstlisting}
The complete type specification for this domain can be found in~\cite{github}.   
The ADE dataset has only two types of entities and thus no type specification is added.

\smallskip \noindent
\emph{Data processing.} 
We preprocessed each raw dataset to extract full sentences and paragraphs for LLM input, rather than tokenized word lists. Our LLM request also specifies a human-readable output, rather than a list of indices or entity spans.

\smallskip \noindent \emph{Baselines for comparison.} Two competitors, (i) {\bf ASPER}~ by \cite{DBLP:journals/tplp/LeCS23} and (ii) {\bf ITER} by ~\cite{hennen-etal-2024-iter}, were chosen as baselines to compare with our proposed method. ASPER utilizes ASP to improve its quality of prediction and ITER has shown to outperform most other joint ER extraction techniques. 
We also implemented a variation of our workflow by replacing the ChatGPT LLM with an ensemble of LLMs, as ensembles generally yield better results than individual models. The ensemble consists of two LLM agents: the fine-tuned ChatGPT and a Gemini agent. Both agents are tasked with auditing the results, and if they both agree on an entity $e$ of type $t$, that entity is included in the output. In reporting the results of this study, we refer to the ensemble of LLMs as {\bf Ensemble}, and the ensemble with the ASP consistency checker as {\bf Ensemble + ASP}.
In all the result tables, we use {\bf E} to represent entity and {\bf ER} to represent entity-relationship. 

\smallskip \noindent \emph{Evaluation metrics.}
We use $F_1$-micro and $F_1$-macro scores to evaluate the model's performance on entities (NER) and entity-relation (ER) tasks. 
$F_1$-micro is calculated using the total true positives, false negatives and false positives. 
$F_1$-macro is the unweighted average of each class type's $F_1$ score. The formula for $F_1$ is $\frac{2TP}{2TP+FP+FN}$. It is generally accepted that systems with better $F_1$-macro score are considered ``better.''

\subsubsection{Default Setting of The LLM+ASP Workflow} 
 
By default, we used a fine-tuned GPT agent, the gpt-4o-2024-08-06 model~\cite{openai}, for the JERE outputs with the ASP consistency checker.

The prompt utilizes one-shot prompt. Each dataset's prompt was specific to that dataset by using the annotation guidelines given in the corpus' accompanying papers. 
For the fine-tuning step, we simulate a low-resource setting. Each dataset is originally split into training (65\%), validation(15\%) and test sets(20\%). 
We randomly selected 10\% of the original training data and 10\% of the original validation data to fine-tune the model, using them for training and validation respectively.
Each fine-tuned model was specific to its dataset. The hyper-parameters were consistent across all datasets, with 5 epochs, a batch size of 1, and a learning rate multiplier of 2. 

The ASP consistency checker uses the ASP program, $\Pi_C$, detailed in the previous section that is also independent from the domain (code see \cite{github}). 
Domain specific information in the form of type specification is provided as an optional input to this program. \\

\begin{table*}[htb]
\centering{
  \begin{tabular}{|c||c|c|c|c||c|c|c|c|}\cline{1-9}
    \multirow{2}{*}{\textbf{Dataset}} &
      \multicolumn{4}{c||}{\textbf{One-shot prompt}} &
      \multicolumn{4}{c|}{\textbf{Fine-tuned GPT}}\\
      \cline{2-9}
    & \multicolumn{2}{c|}{\textbf{F1-Micro}} &
      \multicolumn{2}{c||}{\textbf{F1-Macro}} &
      \multicolumn{2}{c|}{\textbf{F1-Micro}} &
      \multicolumn{2}{c|}{\textbf{F1-Macro}} \\
      \cline{2-9}
    & E & ER & E & ER & E & ER & E & ER \\\cline{1-9}
    CoNLL04 & 
    73.29 & 44.78 & 67.42 & 48.42 &
    80.27 & 58.82 & 74.59 & 57.31 \\
    SciErc & 
    42.83 & 7.89 & 35.56 & 7.37 &
    61.70 & 26.55 & 60.94 & 22.94 \\
    ADE & 
    88.30 & 37.28 & 88.75 & 37.28 &
    90.32 & 82.84 & 90.84 & 82.84 \\\cline{1-9}
  \end{tabular}
  \caption{One-shot prompt vs. Fine-tuned (E: entity; ER: entity-relationship; No ASP Checking)
  }
  \label{table:prompt}
}
\end{table*}

Given that the fine-tuned models with the randomly chosen 10\% of training data consistently outperforms the one-shot prompt 
(Table~\ref{table:prompt}), we used the fine-tuned models throughout the rest of this paper. Additionally, because the models do not provide deterministic responses and may produce hallucinations, we run each model three times to obtain a more robust assessment and report the averaged results.

\subsubsection{Training Time and Model Sizes}
Most of the computational load of our proposed method is handled by OpenAI's servers. Fine-tuning GPT-4o on 400 training samples for the ADE dataset for 5 epochs takes $\approx$15 minutes, with evaluation of the full test set taking an additional 15 minutes. Similar running time is observed on the other datasets. The computation of the ASP consistency checker is efficient and nearly negligible, requiring only $\sim$10 milliseconds to process all predictions per dataset.

Regarding scalability, the main computational cost lies in fine-tuning and prediction. Fine-tuning scales linearly with data size and the number of epochs, while prediction scales linearly with the number of words, as it operates at the sentence level. The ASP consistency checker adds negligible overhead.

We would also report the sizes of the model utilized by our approach and the baselines. Our approach is based on a GPT agent gpt-4o-2024-08-06 model, which has approximately 1.76 trillion parameters. For comparison, the ASPER model uses around 110 million fixed (pretrained) parameters and approximately 20,000 trainable parameters across all datasets. The ITER model has a total of 393 million parameters for all datasets.
The model sizes show one limitation of our approach in that it utilizes larger models compared with the baseline.

\subsection{LLM + ASP vs. State-of-the-Art Systems}
This section shows the effectiveness of our proposed LLM + ASP workflow using the default 
setting stated in Section 4.1.1 
when compared with other baselines. 
The training data that is used in LLM fine-tuning is randomly chosen 10\% of the original training set (default setting). 
For fair comparison, for both ASPER and ITER, we also used 10\% of the original training data. For ITER, the 10\% training data is the same as that used to fine-tune the LLM model. For ASPER, we use the authors' chosen 10\% data to be consistent with their configuration.


\begin{table}[ht]
\centering{
  \begin{tabular}{|@{}c|@{}c|c|c|c|@{}c|c|c|c|@{}c|c|c|c@{}|}
    \cline{1-13}
    \multirow{2}{*}{
    \textbf{Method}} &
      \multicolumn{4}{c|}{\textbf{CoNLL04}} &
      \multicolumn{4}{c|}{\textbf{SciErc}} & 
      \multicolumn{4}{c|}{\textbf{ADE}}
      \\
      \cline{2-13}
    & \multicolumn{2}{c|}{\textbf{F1-Micro}} &
      \multicolumn{2}{c|}{\textbf{F1-Macro}} &
      \multicolumn{2}{c|}{\textbf{F1-Micro}} &
      \multicolumn{2}{c|}{\textbf{F1-Macro}} &
      \multicolumn{2}{c|}{\textbf{F1-Micro}} &
      \multicolumn{2}{c|}{\textbf{F1-Macro}} \\
      \cline{2-13}
    & E & ER & E & ER 
    & E & ER & E & ER
    & E & ER & E & ER\\\cline{1-13}
    GPT+ASP & 
    80.45 & 60.51 & 74.79 & \textbf{58.91}  &
    \textbf{62.32} & \textbf{38.23} & 61.55 & \textbf{35.37} &
    \textbf{90.40} & \textbf{83.89} & \textbf{90.91} & \textbf{83.89} \\
    \cline{1-13}
    Ens.+ASP&
    80.29 & \textbf{60.54} & 74.44 & 58.59 &
    \textbf{62.32} & 37.23 & \textbf{61.64} & \textbf{35.37} &
    89.53 & 82.21 & 90.08 & 82.21 \\ 
    \cline{1-13}
    ASPER & 
    \textbf{81.25} & 52.41 & \textbf{75.90} & 53.27 &
    60.34 & 21.73 & 59.10 & 16.06 &
    86.60 & 75.30 & 86.93 & 75.30\\
    \cline{1-13}
    ITER&
    70.81 & 34.37 & 63.15 & 27.58 &
    56.07 & 10.53 & 55.46 & 10.00 &
    86.49 & 75.70 & 87.10 & 75.70 \\
    \cline{1-13}
  \end{tabular}
  \caption{Performance comparison of different systems (E: entity, ER: entity-relationship)
  }
  \label{table:comparison}
}
\end{table}

Table~\ref{table:comparison} shows the overall results. Boldface numbers indicated systems with the best score in the corresponding category. 
As can be seen, our workflow is comparable to the state-of-the-art supervised models in ER. It consistently outperforms ASPER by \cite{DBLP:journals/tplp/LeCS23} in the ER task. On the SciErc dataset, it excels over ASPER by 20\% raw score where  GPT+ASP can achieve 35.37\% $F_1$-macro while ASPER is at 16.06\% $F_1$-macro.  

Notably, existing state-of-the-art methods perform poorly on the SciERC dataset. Surprisingly, our workflow outperforms ITER by 
more than 25\% raw score.  We attribute this improvement to the ASP consistency checker, which reduces FP and, as a result, enhances the quality of entity-relation resolutions. We want to note that when trained on 100\% of the training data, ITER outperforms our GPT+ASP, that used only the randomly chosen 10\% of the original training data, by 7\% raw score.  

\subsection{Ablation Studies}
This set of experiments is to examine the effect of two components (1) the ASP consistency checker and (2) the ensemble of the LLMs. \\

\noindent 



\begin{table*}[h]
\centering{
    \begin{tabular}{|@{}c@{}|@{}c|c|c|c|@{}c|c|c|c|@{}c|c|c|c|}
    \cline{1-13}
    \multirow{2}{*}{
    \textbf{Methods}} &
      \multicolumn{4}{c|}{\textbf{CoNLL04}} &
      \multicolumn{4}{c|}{\textbf{SciErc}} & 
      \multicolumn{4}{c|}{\textbf{ADE}} 
      \\
      \cline{2-13}
    & \multicolumn{2}{c|}{\textbf{F1-Micro}} &
      \multicolumn{2}{c|}{\textbf{F1-Macro}} &
      \multicolumn{2}{c|}{\textbf{F1-Micro}} &
      \multicolumn{2}{c|}{\textbf{F1-Macro}} &
      \multicolumn{2}{c|}{\textbf{F1-Micro}} &
      \multicolumn{2}{c|}{\textbf{F1-Macro}} \\
      \cline{2-13}
    & E & ER & E & ER 
    & E & ER & E & ER
    & E & ER & E & ER
    \\\cline{1-13}
    GPT+ASP\: &
    80.45 & 60.51 & 74.79 & 58.91  &
    62.32 & 38.23 & 61.55 & \textbf{35.37}  &
    90.40 & 83.89 & 90.91 & 83.89  \\ \cline{1-13}
    GPT&
    80.27 & 58.82 & 74.59 & 57.31 &
    61.70 & 26.55 & 60.94 & \textbf{22.94}  & 
    90.32 & 82.84 & 90.84 & 82.84\\\cline{1-13}
    Ens.+ASP& 
    80.29 & 60.54 & 74.44 & 58.59 &
    62.32 & 37.23 & 61.64 & \textbf{35.37} &
    89.53 & 82.21 & 90.08 & 82.21 \\\cline{1-13}
    Ensemble & 
    80.51 & 58.15 & 75.07 & 56.75  &
    61.20 & 26.14 & 60.59 & \textbf{25.15}  &
    90.10 & 82.06 & 60.41 & 82.06  \\\cline{1-13}
    \end{tabular}
    \caption{Results for ablation study (E: entity, ER: entity-relationship); Ens.: GPT and Gemini Ensemble}
    \label{table:ablation}
}
\end{table*}

\smallskip \noindent \emph{Effect of ASP consistency checker.} 
The first experiment demonstrates the contribution of the ASP consistency checker to our workflow (see Table~\ref{table:ablation}). Boldface numbers highlight the scores in SciERC dataset, the most difficult dataset for JERE. 
We compare the outputs (entities and relationships) from our default workflow, GPT+ASP (Row 1, Table ~\ref{table:ablation}), with those from the fine-tuned GPT alone (Row 2, Table~\ref{table:ablation}), as well as the results from Ens.+ASP (Row 3) and the ensemble model alone (Row 4). 
As can be observed, both the F1-macro and F1-micro scores with the ASP consistency checker (Rows 1 and 3) 
improve upon the corresponding version without the ASP consistency checker (Rows 2 and 4) for ER, sometimes more than 30\% (SciErc dataset). 
This demonstrates the effectiveness of the ASP checker in the process when domain specifications are available. In ADE, we do not see a large increase in the ER scores since there is only one relationship type to extract and thus less to reduce based on the domain knowledge. \\


\begin{table*}[htp]
\centering{
  \begin{tabular}{|c|c|c|c|c|c|c|c|c|c|c|c|c|}
    \cline{1-13}
    \multirow{3}{*}{\textbf{Dataset}} &
      \multicolumn{6}{c|}{\textbf{GPT}} &
      \multicolumn{6}{c|}{\textbf{GPT+ASP}}\\
      \cline{2-13}
      \multicolumn{1}{|c|}{} &
      \multicolumn{3}{c|}{\textbf{E}} &
      \multicolumn{3}{c|}{\textbf{ER}} &
      \multicolumn{3}{c|}{\textbf{E}} &
      \multicolumn{3}{c|}{\textbf{ER}}\\
      \cline{2-13}
    & TP & FP & FN & TP & FP & FN & TP & FP & FN & TP & FP & FN\\
    \cline{1-13}
    CoNLL04 &
    881 & 258 & 176 & 262 & 224 & 144  &
    883 & 256 & 174 & 264 & 203 & 142  \\
    \cline{1-13}
    SciErc &
    1003 & 575 & 670 & 244 & 713 & 639 & 
    1004 & 574 & 640 & 339 & 482 & 614 \\
    \cline{1-13}
    ADE & 
    991 & 98 & 114 & 571 & 112 & 125 &
    992 & 98 & 113 & 579 & 105 & 117 \\
    \cline{1-13}
  \end{tabular}
  \caption{Effect of ASP to improve FP (Randomly chosen 10\% Training Data). (TP: True Positive, FP: False Positive, FN: False Negative; E: entity, ER: entity-relationship)} 
  \label{table:tpfp}
}
\end{table*}

We conducted a more detailed analysis of this improvement by examining how many entities and entity relationships are truly or falsely reported as positive. Table~\ref{table:tpfp} presents the numbers for GPT+ASP and GPT alone. The results show that, across all three datasets, the number of falsely reported positive entity-relationships are reduced with the use of the ASP consistency checker. Datasets with more type-specifications, like SciErc, benefited most from the consistency checker - going from 713 FP values to 482.


\smallskip \noindent \emph{ Effect of LLM ensemble. } The second experiment examines whether the LLM ensemble helps improve an individual LLM agent. The result is reported in Table~\ref{table:ablation}. As it turns out, the ensemble, in its current use, does not perform better than the single GPT agent, with or without the ASP checker. This can be seen in the results in Row 2 vs. Row 4 (GPT vs. Ensemble) and Row 1 vs. Row 3 (GPT + ASP vs. Ensemble + ASP). The reason for this reduced performance is that TP entities, detected by GPT, are removed from consideration which, ultimately, reduces the $F_1$-macro/micro scores.

\subsection{Effect of Amount of Training Data}
Table \ref{table:trainingdiff} shows the results of our default workflow with different versions of GPT, fine-tuned  on 5\%, 10\%, and 15\% of training data, respectively. These small percentages of training data are all randomly chosen. For each setting, the LLM model is fine-tuned three times and the reported number is the average of the results from the three fine tuned models. 



\begin{table*}[htb]
\centering{
  \begin{tabular}{|c||c|c|c|c||c|c|c|c||c|c|c|c|}
    \cline{1-13}
    \multirow{2}{*}{
    \textbf{}} &
      \multicolumn{4}{c||}{\textbf{5\% TD+ ASP checker}} &
      \multicolumn{4}{|c||}{\textbf{10\% TD + ASP checker}} & 
      \multicolumn{4}{c|}{\textbf{15\% TD + ASP checker}}
      \\
      \cline{2-13}
    & \multicolumn{2}{c|}{\textbf{F1-Micro}} &
      \multicolumn{2}{c||}{\textbf{F1-Macro}} &
      \multicolumn{2}{c|}{\textbf{F1-Micro}} &
      \multicolumn{2}{c||}{\textbf{F1-Macro}} &
      \multicolumn{2}{c|}{\textbf{F1-Micro}} &
      \multicolumn{2}{c|}{\textbf{F1-Macro}} \\
      \cline{2-13}
    & E & ER & E & ER 
    & E & ER & E & ER
    & E & ER & E & ER\\
    \cline{1-13}
    C &  77.68 & 58.07 & 72.52 & 58.64  
    & 80.45 & 60.51 & 74.79 & 58.91  
    & 80.20 & 57.71 & 73.58 & 55.31 \\
    \cline{1-13}
    S & 59.14 & 32.46 & 59.15 & 31.14 &
     62.32 & 38.23 & 61.55 & 35.37 &
     64.16 & 40.63 & 64.10 & 35.05 \\
    \cline{1-13}
    A & 90.13 & 79.61 & 90.61 & 79.61  &
     90.40 & 83.89 & 90.91 & 83.89  &
     90.73 & 84.00 & 91.16 & 84.00 \\
    \cline{1-13}
  \end{tabular}
  \caption{Results different percentages of training data on fine-tuned ChatGPT model. (E: entity; ER: entity-relationship; TD: Training Data; C: CoNLL04; S: SciErc; A: ADE)}
  \label{table:trainingdiff}
}
\end{table*}

Overall, the workflow performs better with more data with some exception. Its performance seems to be domain-dependent. We can observe distinct improvement from 5\% to 10\% from for each dataset. However, in the ADE results we can see improvements only in the ER task - with there being minimal difference between the models fine-tuned on 10\% and 15\%. The overall scores are better for SciErc with the exception of the $F_1$-macro score for the ER task between the 15\% and 10\% models. In the CoNLL04 dataset, we see a reduced score when comparing 10\% and 15\% data.

\section{Conclusion}\label{sec:conclusion}

In this paper, we propose a generic workflow for joint entity-relation extraction using LLMs and ASP. The workflow is used to perform the JERE task on arbitrary domains. Due to the LLM's capability in natural language understanding, our system can perform the JERE task on \emph{unannotated text}, which sets it apart from contemporary systems that require large amounts of annotated tokenized text. The workflow can exploit domain-specific information, when available, to improve its performance. In addition, our approach offers greater flexibility and scalability, as it can adapt to new domains with minimal additional 
fine-tuning. We demonstrate the usefulness of the proposed workflow through experiments with limited training data on three well-known benchmarks for JERE. The results of our experiments show that the LLM+ASP workflow is better than state-of-the-art JERE systems in several categories. In the near future, we plan to explore using this workflow to extract knowledge graphs as they consist of entities and relations.

\section*{Acknowledgment}
This work has been supported by NSF award \#1914635. 
The first and last authors were also supported by 
NRC Grant 31310022M0038.

\bibliographystyle{eptcsalpha}
\bibliography{references}
\end{document}